# When More Data Is Not Enough

## The Context-Sufficiency Frontier in Generative AI Personalization

**Merieme Askour** and **Ayoub Merimi**

**Abstract**

Personalization has long relied on customer data to infer what an individual is likely to value. We call this **customer evidence**: the customer's historical behavior and preferences. Generative AI extends personalization by allowing providers to supply changing situational information at the moment a response is produced, without encoding every condition in advance. We define this **provider-side context** as information about what is possible, permitted, or advisable now. This flexibility creates a new problem: once context becomes easy to supply, more is not necessarily better. We develop a theory of **context sufficiency** in which the relevance of context to the customer's current intent matters more than its volume. The theory identifies four states, insufficiency, sufficiency, saturation, and interference, and introduces the **Context-Sufficiency Frontier** to locate the minimal relevant set. In a full-factorial experiment with a generative recommender at a large home-furnishing retailer, relevant context improved appropriateness, while irrelevant context reduced it and destabilized retrieval. The framework shifts personalization from supplying more context toward identifying what the current interaction actually requires and enforcing constraints throughout the service process.

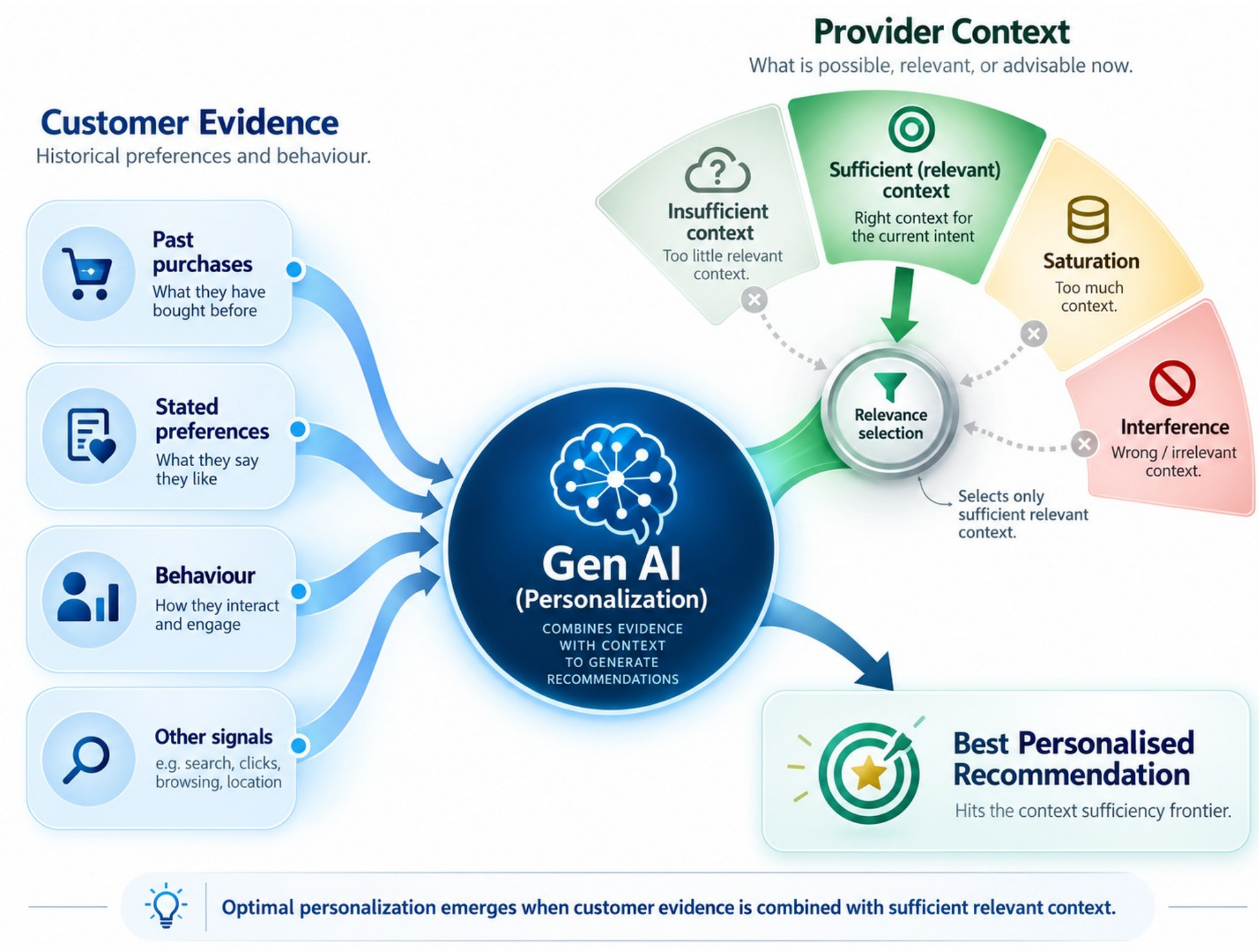

**Graphical abstract.** *Optimal personalization combines customer evidence with the provider-side context the current request actually needs.*

## 1 Introduction

Personalization rests on a familiar premise: the more a provider knows about a customer, the better it can tailor a service response (Chandra et al. 2022; Rafieian and Yoganarasimhan 2023). We call this knowledge **customer evidence**: the customer's history, behavior, and preferences, which indicate what the customer is likely to value. Yet customer evidence alone cannot determine what a provider should do in a specific interaction. A product can fit a customer's preferences and still be the wrong recommendation now because it cannot be fulfilled, does not fit the immediate situation, conflicts with a provider constraint, or has become inappropriate after an external event. The relevant criterion is therefore appropriateness in the interaction, not predicted preference alone.

We call the missing input **provider-side context**. We use context narrowly: situational information held by the provider that changes what constitutes an appropriate response to the customer's current request. A supply disruption may make an otherwise suitable product unavailable; an unusual competitor move may change what should be emphasized; a new operational constraint may rule out an option that customer evidence would otherwise favor. The same fact can therefore be decisive in one interaction and irrelevant in the next. The problem is not whether context matters in general, but which context matters now.

We distinguish five types of provider-side context, **environment**, **goal**, **operational**, **world**, and **guardrail** context, each defined by the decision question it answers. Their value is instance-specific: a signal matters only when it bears on the customer's current request.

Generative AI makes this problem more consequential because provider-side context can be supplied when a personalized response is generated, without permanently engineering every condition into the underlying recommendation system (Zhao et al. 2024; Grewal et al. 2025). That flexibility also creates a failure mode. If context is easy to provide, firms may provide too much. Additional signals can be relevant, redundant, irrelevant, or conflicting. More context may therefore improve a response, do nothing, or make it worse (Shi et al. 2023; Liu et al. 2024). The problem is central to service research. Frontline employees have long personalized service by reading the situation and adapting to it (Gwinner et al. 2005), and service systems have long substituted stored customer and contextual information for that interaction (Glushko and Nomorosa 2013). As AI takes over service tasks (Huang and Rust 2018, 2021b) and generative AI permeates frontline routines (Reinhard et al. 2026), this situational judgment must be transferred into the generative system as provider-side context, and managing that transfer is part of managing AI-delivered service (Ostrom et al. 2021). The question this paper addresses is therefore which provider-side context a personalization decision requires, when it has received enough, what happens when more is supplied, and what is lost when the selection of that context is delegated to the model.

We develop a framework and theory of **context sufficiency** to explain when additional context helps, stops helping, or harms. The central claim is that relevance to the request matters more than context volume. As signals accumulate, a personalization instance can move through four states: insufficiency, when relevant information is missing; sufficiency, when the minimal relevant set is present; saturation, when additional signals add little; and interference, when irrelevant or conflicting signals reduce appropriateness. The **Context-Sufficiency Frontier** identifies the highest appropriateness attainable with a given number of context signals and the smallest set required to reach it.

We test the theory in a full-factorial offline experiment using the catalog of a large home-furnishing retailer, supplying five context signals in all 32 combinations to a generative recommender operating around the retailer's catalog search. Across the 320 runs of the design (8,000 ranked recommendation positions), relevant signals improved appropriateness while irrelevant signals reduced it. Beyond the relevant signals, additional context made recommendations less consistent, mainly by destabilizing retrieval, and one provider-goal signal carried products into the final recommendation that the service rules were intended to exclude.

The study contributes a decision-level account of personalization that separates customer evidence from provider-side context and explains when additional context helps, stops helping, or interferes. For service theory, it recasts personalization in generative service systems as the transfer of situational judgment from frontline and service-system actors into the system's input, and it shows what is lost when the selection of that context is delegated to the model. It also offers a managerial unit, the decision context package, for deciding which context a personalization decision should receive. The evidence comes from a controlled offline experiment with two constructed customers, one request, and one primary model, so it establishes the relevance mechanism and shows a rise-and-fall pattern consistent with the theory in this setting; the final section sets out the broader tests needed.

## 2 Conceptual Foundations of Context-Sufficient Personalization

To theorize when context is sufficient, we first specify the personalization decision to which sufficiency applies. Personalization is commonly defined as adapting a service offering to an individual customer using what the provider knows about that customer (Chandra et al. 2022). Service research has long separated the information personalization draws on: Glushko and Nomorosa (2013) distinguish customer models, service models, and contextual information used to adapt the service at delivery time. We retain the customer as the beneficiary of personalization but distinguish three inputs to the response: **customer evidence**, which indicates what the customer is likely to value; **current intent**, which identifies what the customer is trying to accomplish in this interaction; and **provider-side context**, which establishes what is possible, permitted, or advisable at the moment of response. **Table 1** defines these constructs and distinguishes them from neighboring concepts.

**Table 1.** Constructs of the Study

| Construct | Definition | Distinguished from |
|---|---|---|
| Personalization instance | One customer, one current request, and the response the provider generates for that request. The unit of analysis. | The customer relationship, which extends across interactions and accumulates evidence over time. |
| Customer evidence | Information about the customer's historical behavior, characteristics, and preferences, including transactions, browsing, searches, stated preferences, saved items, returns, and complaints. Indicates what the customer is likely to value. | Provider-side context, which concerns the circumstances of the decision rather than the customer. |
| Current intent | The objective expressed or implied by the customer's request in the focal interaction. The reference point for appropriateness. | Customer evidence, which indicates what the customer is likely to value rather than what the customer is trying to accomplish now. |
| Provider-side context | Situational information available to the provider that changes what constitutes an appropriate response to the current request. Five types: environment, goal, operational, world, and guardrail. | The model's context window, a technical input capacity; the prompt, which specifies the task the model should perform. |
| Personalization appropriateness | The degree to which a response preserves the customer's current intent, reflects the provider-side context relevant to that intent, and respects applicable constraints. | Predicted preference; recommendation change, which can occur without improvement. |

### 2.1 The Personalization Instance

The unit of analysis is the **personalization instance**: one customer, one current request, and the response the provider generates for that request. This differs from the customer relationship, which extends across interactions and accumulates evidence over time. The distinction matters because context sufficiency is not a property of a customer or of a personalization system in general. It is a property of a particular decision. The same customer may require different provider-side context when furnishing a bedroom, replacing a damaged item, or arranging an urgent delivery. Accordingly, all claims about insufficiency, sufficiency, and interference in this study refer to what a particular personalization instance requires.

### 2.2 Customer Evidence and Current Intent

**Customer evidence** is information about the customer's historical behavior, characteristics, and preferences, including transactions, browsing, searches, stated preferences, saved items, returns, and complaints. It records what happened or what is known about the customer and thereby informs what the customer is likely to value. Recommendation systems use the customer's behavior and the behavior of similar customers for precisely this purpose (Rafieian and Yoganarasimhan 2023), and AI-based personalization similarly treats richer customer knowledge as an important input to tailored service (Huang and Rust 2021a).

Customer evidence is necessary for personalization, but it does not specify the decision by itself. A customer's history may indicate a preference for a particular style, category, or price range without indicating what that customer is trying to accomplish today. We therefore distinguish customer evidence from **current intent**, the objective expressed or implied by the customer's request in the focal interaction. Evidence answers, *What is this customer likely to value*? Intent answers, *What is this customer trying to accomplish now*?

The distinction becomes consequential when the two point in different directions. A customer whose prior purchases indicate a preference for large, dark furniture might now ask to make a small bedroom feel more spacious. Reproducing the historical preference faithfully could produce a highly personalized response that nevertheless fails the request. Current intent therefore provides the reference point for appropriateness. Customer value is realized relative to the customer's goals in the situation of use (Zeithaml et al. 2020), and AI-delivered service ultimately must remain aligned with the customer's intentions rather than the provider's objectives alone (Huang and Rust 2024).

### 2.3 Provider-Side Context

Even customer evidence and current intent together may be insufficient to determine the appropriate response. A recommendation that fits both can still be impossible to fulfill, inadvisable under present conditions, inconsistent with a constraint, or inappropriate because circumstances have changed. We define **provider-side context** as situational information available to the provider that changes what constitutes an appropriate response to the current request.

This definition is functional rather than categorical. A warehouse disruption is an organizational fact, but it becomes a personalization context only when it affects what the provider can appropriately offer in the focal interaction. A competitor promotion is context when it changes the relevant response to the request, but not merely because the provider knows about it. The same fact can therefore be relevant context for one personalization instance and irrelevant information for another. Context is defined by its bearing on the decision, not simply by its availability or source.

Provider-side context is also distinct from two uses of "context" common in generative AI. It is not the model's **context window**, which describes a technical input capacity, and it is not the **prompt**, which specifies the task the model should perform. Provider-side context supplies facts about the circumstances under which that task must be performed.

We distinguish five types of provider-side context, environment, goal, operational, world, and guardrail, which the next section specifies. They do not imply that every personalization instance requires one signal of each type; this instance dependence is central to the theory that follows.

### 2.4 Personalization Appropriateness

We define **personalization appropriateness** as the degree to which a response preserves the customer's current intent, reflects the provider-side context relevant to that intent, and respects applicable constraints. This criterion differs from predicted preference. A recommendation can closely match historical preferences yet be inappropriate for the present request. It also differs from recommendation change. A context signal can alter a recommendation without making it better, particularly in generative systems whose outputs vary across otherwise identical runs (Ouyang et al. 2025).

Appropriateness thus provides the outcome against which context should be evaluated. Relevant context should improve the response because it resolves something the personalization instance needs to know. Context that adds nothing to that decision should not receive the same status merely because it is available. This distinction leads to the theoretical problem developed next: **which provider-side context does a personalization instance require, when has it received enough, and what happens when additional context continues to be supplied?**

## 3 A Theory of Context Sufficiency

Context sufficiency asks a simple question: **how much provider-side context does a generative personalization system need for this interaction?** Our answer is not all the context the provider has. It is the smallest set of context that covers what matters to the customer's current request.

The theory develops this argument in three steps. First, provider-side context can come from different parts of the provider's situation, but only some of those parts will matter in a given interaction. Second, as relevant context is supplied, the interaction moves from insufficiency to sufficiency; beyond that point, additional context can become redundant or interfere with the response. Third, the Context-Sufficiency Frontier identifies where this point occurs.

### 3.1 The Context-Sufficiency Framework

Figure 1 presents the Context-Sufficiency Framework, which combines the three inputs and the appropriateness criterion defined above.

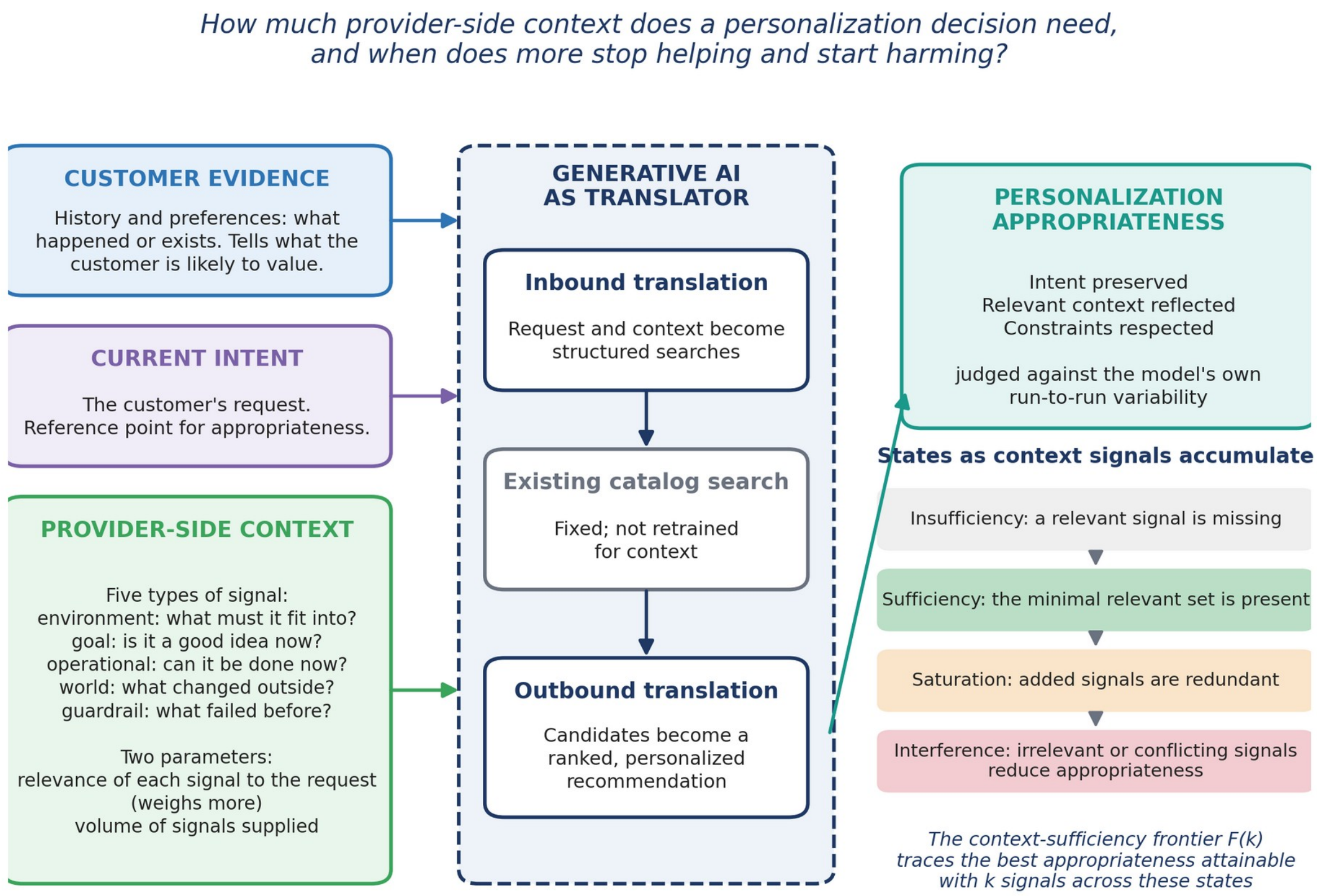

**Figure 1. The Context-Sufficiency Framework.** Customer evidence, current intent, and provider-side context enter a generative AI layer that translates the request and context into structured searches of an existing catalog search and translates the results into a ranked, personalized recommendation. Appropriateness is judged on three criteria. As context signals accumulate, a personalization instance passes through insufficiency, sufficiency, saturation, and interference.

The framework distinguishes five types of provider-side context according to the question each type answers (Figure 2).

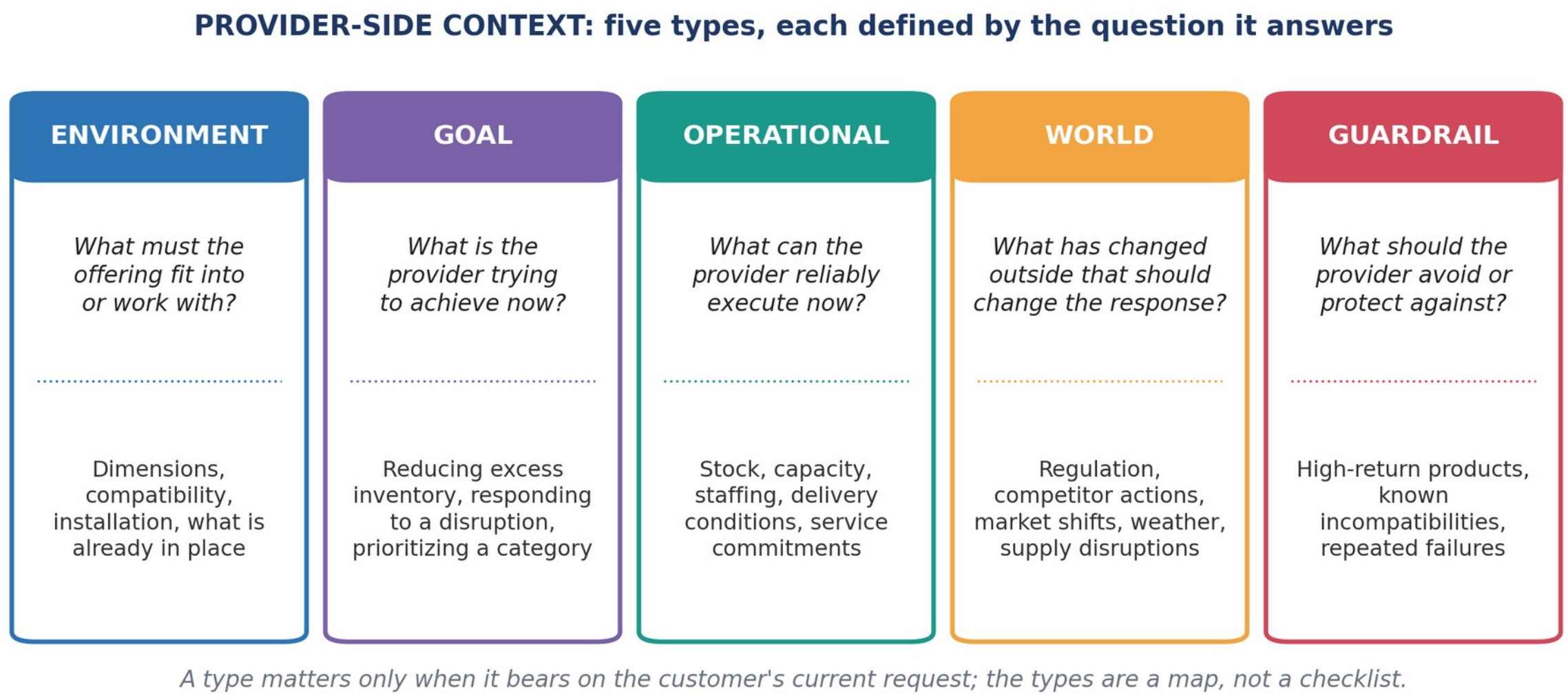

**Figure 2. Five Types of Provider-Side Context.** Each type is defined by the decision question it answers; examples follow the definitions in the text.

**Environment context** asks: *What must the offering fit into or work with?* It covers the existing setting, system, or configuration in which the offering must function, including physical dimensions, technical requirements, compatibility, installation conditions, what is already in place, and relevant dependencies. A recommendation can match the customer's preferences yet still be inappropriate because it does not fit or function within that environment.

**Goal context** asks: *What is the provider trying to achieve now?* It covers current priorities, strategic shifts, targets, crises, or initiatives that may change what the provider should emphasize. Examples include reducing excess inventory, responding to a sudden disruption, changing strategic direction, protecting a critical business objective, or prioritizing a product category. Goal context captures what the provider is actively trying to accomplish at that moment and is relevant only when that objective can be pursued without undermining the customer's current intent.

**Operational context** asks: *What can the provider reliably execute now?* It covers the current resources, staffing, capacity, service commitments, process constraints, availability, delivery conditions, and quality levels that determine what the provider can actually perform or promise. A response can be desirable in principle yet inappropriate if the provider lacks the capacity to deliver it reliably or at the required quality.

**World context** asks: *What has changed outside the provider that should change the response?* It covers external developments such as government policy, regulation, economic conditions, competitor actions, market shifts, weather conditions, supply disruptions, or other events that make previously valid information incomplete or unsafe to use without adjustment. World context matters when external change makes previously valid information, assumptions, or recommendations no longer suitable.

**Guardrail context** asks: *What should the provider avoid or protect against?* It covers known risks, recurring failures, undesirable outcomes, policy limits, and lessons from prior experience that should constrain the response. Examples include high-return products, known incompatibilities, repeated service failures, safety concerns, or combinations associated with poor outcomes. Guardrail context matters when an option may appear suitable but carries a known reason not to recommend, promise, or pursue it.

These five types are not a checklist that every interaction must satisfy. They provide a map of where relevant provider-side context may come from. One request may require operational context but no goal context; another may require environment and world context together. Several signals may also address the same issue. What matters is therefore not whether all five types are represented, but whether the context supplied **covers what matters to the interaction**.

This leads to a central distinction in the theory. **Relevance concerns whether a signal bears on the current interaction. Coverage concerns whether the supplied set adequately represents the circumstances that matter. Volume is simply how many signals are supplied.** A larger context set can therefore provide less useful coverage than a smaller, better-targeted one.

No context type is inherently more important than another. Its importance depends on the request. Stock availability may be critical when immediate delivery matters and irrelevant when the request concerns style inspiration. A competitor action may matter in one category and have no bearing in another. Goal context may help when it can be pursued within the customer's intent, but become inappropriate when it redirects the response toward a provider priority the request does not support.

This instance dependence is what makes context sufficiency possible. Providers may hold hundreds of situational facts, but only a subset will matter to the response being produced. The theoretical problem is therefore not how much context the provider has, but **which context the interaction requires and whether that requirement has been sufficiently covered**.

### 3.2 Relevance and Coverage

A context signal is **relevant** when it bears on what constitutes an appropriate response to the customer's current request. Relevance is therefore a relationship between the signal and the interaction, not a permanent property of the signal itself.

Consider a sudden signal indicating extreme heat. It may be highly relevant to a request involving outdoor products or temperature-sensitive goods, but irrelevant to a request for an office desk. The information has not changed. What changes is whether it matters to the response.

Relevance also does not mean that every relevant signal must be supplied. Two signals can concern the same part of the request and therefore provide overlapping information. The customer's request or customer evidence may already contain part of what a context signal would add. A signal about a trend toward mirrors, for example, may be relevant to a request to make a room feel larger, while adding little because the request already points toward mirrors. This distinction resolves an important issue. **Context sufficiency does not mean supplying every relevant signal. It means supplying enough context to cover the relevant** **circumstances.** Once an issue has already been adequately covered, another signal that says substantially the same thing is redundant.

This view is consistent with the broader principle that information has value when it resolves uncertainty relevant to the task at hand (Galbraith 1974). Context-aware recommendation similarly gains from situational variables when those variables bear on the recommendation being produced (Adomavicius et al. 2022; Kulkarni and Rodd 2020). The amount of information alone says little about its value.

> **Proposition 1 (Relevance):** A *provider-side context signal improves personalization appropriateness when it supplies information relevant to the current interaction that is not already adequately covered. Irrelevant context does not improve appropriateness and may reduce it when the generative system acts on it.*

### 3.3 Why Generative AI Changes the Problem

Context has long been used in recommendation systems. The difference with generative AI is how easily new context can enter the system (Figure 3).

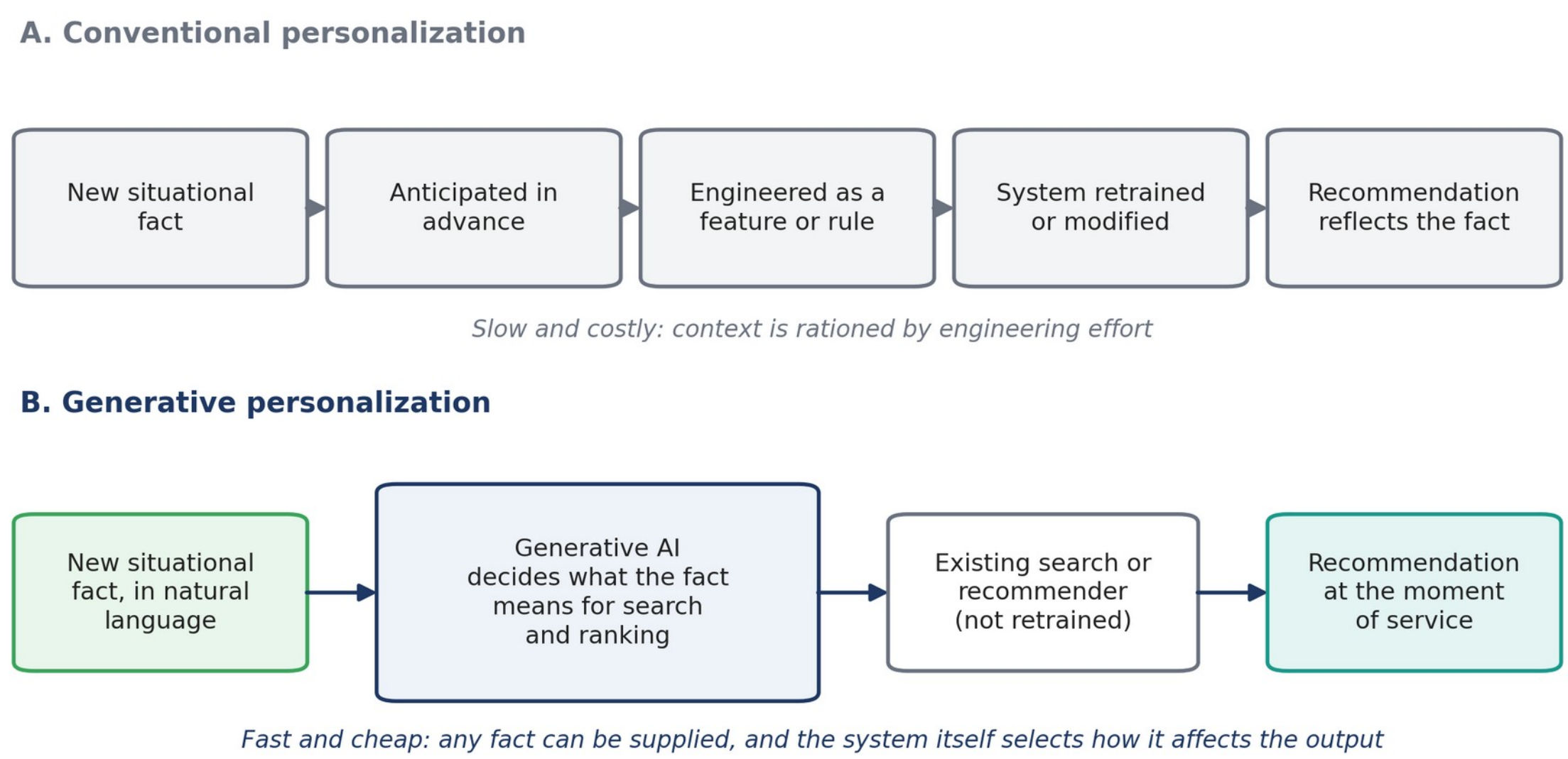

**Figure 3. Why Generative AI Changes the Problem.** In conventional systems, a new situational fact must be anticipated and engineered into the system before it can affect recommendations. In generative personalization, the fact can be supplied in natural language at the moment of service and translated around an existing system, which removes the engineering limit on how much context is supplied.

In conventional systems, contextual variables normally have to be anticipated, represented as features or rules, and engineered into the system before they can affect recommendations (Adomavicius et al. 2022; Kulkarni and Rodd 2020). Generative AI allows providers to supply changing information directly when a response is produced. A supply disruption, competitor action, temporary priority, or newly discovered failure pattern can be provided in natural language without retraining the underlying recommendation system.

This flexibility makes provider-side context much easier to use. It also removes a practical limit that previously restricted how much context could be added. If almost any situational fact can be supplied, a natural response is to give the generative system everything that might matter.

The theory argues that this can be counterproductive because a generative system does not simply receive context and apply only the parts the provider intended. It selects for itself how the supplied information affects its output. It may ignore a signal, use it appropriately, combine it with other signals, or allow it to influence parts of the response that the provider did not intend.

Existing evidence supports this concern. Language models can be distracted by irrelevant context (Shi et al. 2023), use relevant information less reliably as inputs become longer or more crowded (Liu et al. 2024; Levy, Jacoby, and Goldberg 2024), respond to features that should not matter to the task (Sclar et al. 2024), and handle conflicting information inconsistently (Xu et al. 2024). Supplying unnecessary context is therefore not always neutral.

This property of generative AI is central to the theory. Relevant context creates an opportunity to improve the response. Irrelevant or conflicting context creates an opportunity for the system to act on information it did not need.

### 3.4 Four States

As provider-side context is supplied, a personalization instance can move through four states: insufficiency, sufficiency, saturation, and interference. These states describe what the supplied context contributes to the interaction, not different types of context. Figure 4 illustrates the progression.

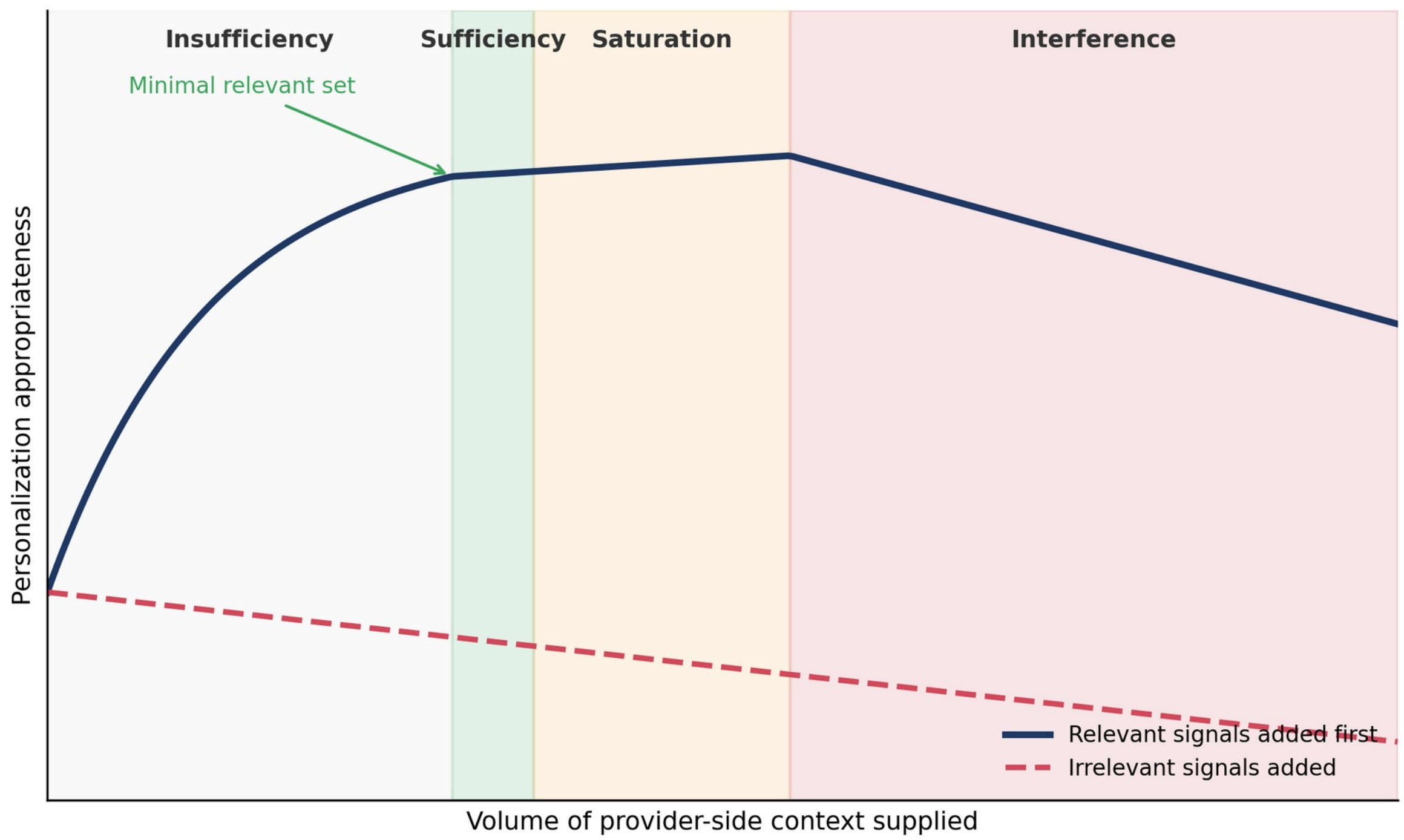

**Figure 4. The Four States of Context Sufficiency.** The horizontal axis is the volume of provider-side context supplied; the vertical axis is personalization appropriateness. The solid line traces appropriateness when relevant signals are added first: it rises through insufficiency, reaches sufficiency at the minimal relevant set, flattens in saturation as further signals become redundant, and declines in interference. The dashed line traces appropriateness when irrelevant signals are added. Conceptual illustration; no data.

**Context insufficiency** occurs when information relevant to the interaction is still missing. The system may understand the customer's preferences and current request while lacking information needed to respond appropriately. It may recommend a product that cannot be delivered, does not fit the environment, or has become unsuitable because circumstances changed. Adding a context signal that resolves such a missing issue should improve appropriateness.

**Context sufficiency** occurs when the supplied context contains the minimum information needed to address the relevant circumstances of the interaction. Nothing consequential remains uncovered, and no additional signal is required to produce the best attainable response. Sufficiency therefore concerns the smallest adequate set, not the largest available set.

The sufficient set need not represent every type of provider-side context. It may contain one type or several, depending on what the interaction requires. It can also differ across interactions involving the same provider or customer because current intent determines which circumstances are relevant.

> **Proposition 2 (Sufficiency):** *For a personalization instance, the highest attainable appropriateness can be reached with a subset of the available provider-side context. The sufficient set is the smallest subset that adequately covers the context relevant to that interaction.*

**Context saturation** occurs when additional context no longer provides meaningful new information. A signal may still concern the interaction, but what it contributes is already represented by the customer's current intent, customer evidence, or other supplied context. Its marginal contribution is therefore small or zero.

Saturation is a property of redundancy, not volume itself. A sixth signal can remain valuable if it resolves something the first five did not. Conversely, the first context signal added can already be redundant if the information it contains is evident from the customer's request or evidence.

> **Proposition 3 (Saturation):** *The marginal contribution of a context signal declines to the extent that its information is already covered by current intent, customer evidence, or other supplied context.*

**Context interference** is the fourth state. It occurs when additional context does more than fail to help and instead reduces personalization appropriateness. Once the circumstances relevant to the interaction have been sufficiently covered, additional signals may be irrelevant, redundant, or conflicting. A generative system may ignore such signals, but it may also act on them. When it does, unnecessary context can compete with relevant information, alter what the system retrieves, change how signals are reconciled, or give an otherwise secondary consideration too much influence over the response.

Interference does not imply that any particular type of provider-side context is inherently harmful. Environment, goal, operational, world, and guardrail context can each improve personalization when relevant to the interaction. The problem arises when a signal is supplied without contributing information the interaction requires, or when it conflicts with information that should govern the response. For example, a provider goal such as reducing excess inventory can appropriately distinguish among products that already satisfy the customer's request. The same goal becomes problematic if the system uses it to surface products that do not fit the request or violate an applicable constraint. Generative systems are expected to serve customer intentions rather than the firm's interests alone (Huang and Rust 2024), and customers can react negatively when AI-driven personalization appears to privilege the provider at their expense (Puntoni et al. 2021).

Interference in generative AI is also distinct from human information overload. Human overload is commonly attributed to limits in attention or cognitive processing (Eppler and Mengis 2004; Chernev, Böckenholt, and Goodman 2015; Thompson, Hamilton, and Rust 2005). A generative system does not become tired as more information is supplied. Rather, additional input can change what the system attends to, retrieves, and generates. Context that the interaction does not require can therefore affect the response even after sufficient relevant information is already present.

Interference may consequently begin before the final recommendation is produced. In a generative recommender, supplied context can affect the searches generated, the products entering the candidate set, and the subsequent ranking of those products. If an unnecessary or conflicting signal changes retrieval, later stages operate on an altered candidate pool and may not be able to recover the options displaced earlier.

### 3.5 Context Interaction

Interaction is not a fifth state. It describes how context signals can change one another's contribution within the four-state progression.

Some signals are complementary. Knowing what a recommendation must fit into can become more useful when the system also knows what is currently available. Environment and operational context may therefore contribute more together than either does alone. Other signals can conflict. A provider goal may favor an option that a guardrail discourages, or a world signal may make an option attractive while an operational constraint makes it infeasible. The contribution of a signal therefore depends partly on the set of information into which it is introduced.

These interactions can affect both appropriateness and stability. Complementary signals can jointly resolve an issue that neither resolves alone. Conflicting signals require the generative system to reconcile competing considerations, and that reconciliation may vary across repeated generations. Context volume by itself therefore cannot explain personalization quality.

> **Proposition 4 (Interaction):** *Context signals that provide complementary information increase one another's contribution to personalization appropriateness, whereas conflicting signals reduce one another's contribution and can reduce response stability.*

> **Proposition 5 (Interference):** *Once context sufficiency has been reached, adding provider-side context that is irrelevant to the interaction or conflicts with the sufficient set reduces personalization appropriateness when the generative system acts on that context and is expected to reduce response stability.*

Not every irrelevant or conflicting signal must produce observable harm in every generation. A generative system may ignore it. The theoretical claim is narrower: once sufficiency has been reached, context that adds nothing required by the interaction creates no additional informational benefit but introduces a pathway through which retrieval, generation, or signal reconciliation can reduce appropriateness.

### 3.6 The Context-Sufficiency Frontier

The four states describe how personalization changes as context accumulates. The **Context-Sufficiency Frontier** provides a way to locate those states.

Let E represent customer evidence, I current intent, and S the set of provider-side context supplied. Personalization appropriateness can be written as

$$A(E,I,S).$$

For a given personalization instance, customer evidence and current intent are fixed while the supplied context varies. We can therefore write appropriateness more simply as

$$A(S).$$

Let $k$ be the number of context signals in S. For the same value of $k$, many different sets can be supplied. Three signals that directly cover what matters to the interaction may perform very differently from three signals that are redundant or irrelevant.

The Context-Sufficiency Frontier is therefore defined as

$$F(k) = \max_{S:|S|=k} A(S).$$

For each number of signals $k$, the frontier identifies the highest appropriateness attainable by any context set of that size (Figure 5). It is a ceiling rather than an average. The difference between the frontier and the average set of the same size captures the value of supplying the right context rather than simply supplying more context.

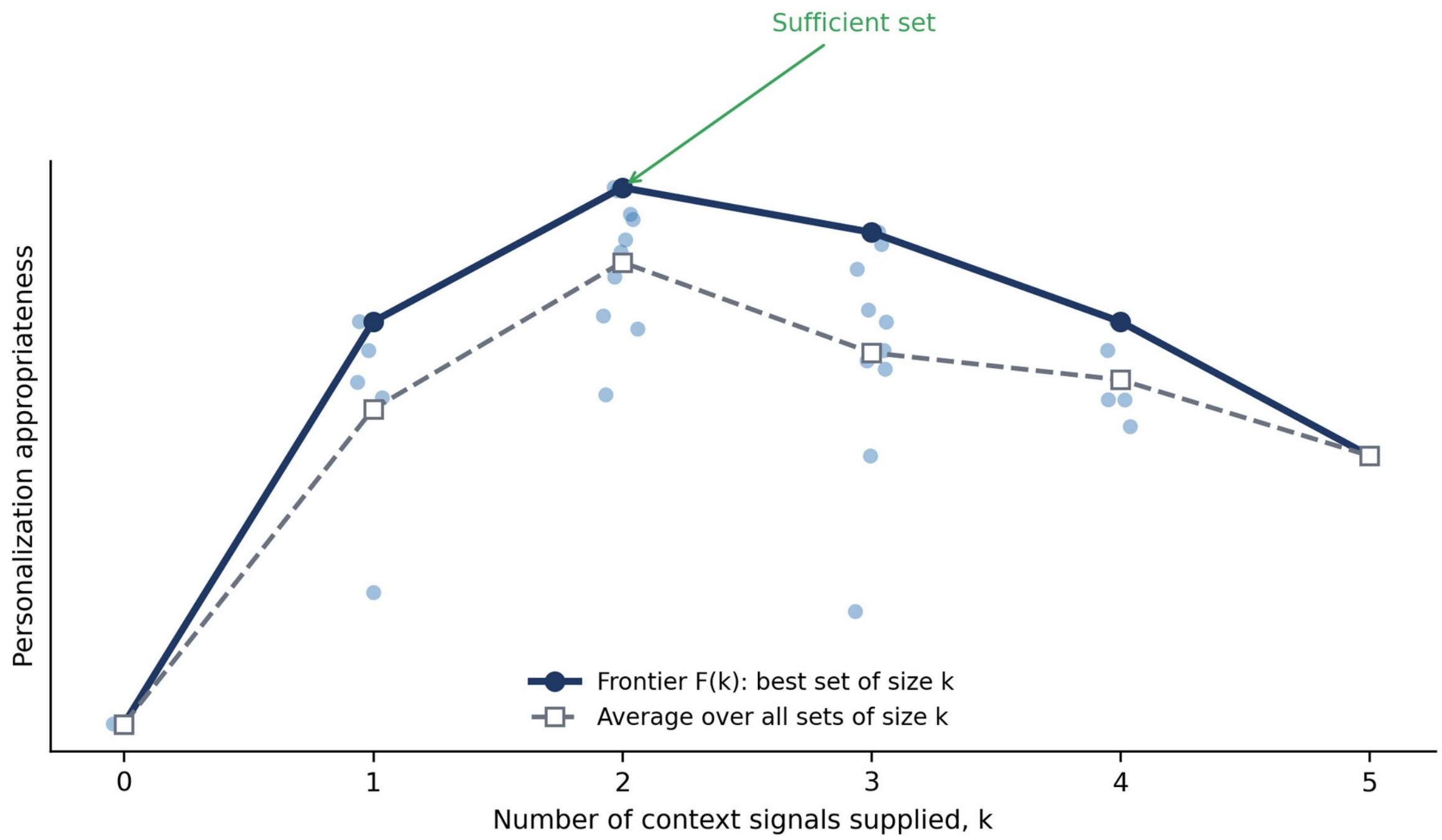

**Figure 5. The Context-Sufficiency Frontier (Illustrative).** Each open circle is one possible set of k context signals; with five signals there are 1, 5, 10, 10, 5, and 1 possible sets at k = 0 to 5. The solid line is the frontier F(k), which connects the best set at each k; the dashed line is the average over all sets of the same size. Conceptual illustration; no data.

The shape of the frontier maps directly onto the four states. While the frontier rises, the interaction remains in **insufficiency** because additional context can still cover something important. **Sufficiency** occurs at the smallest $k$ at which the frontier reaches its highest level, within a tolerance appropriate to the system's own variability. If the frontier then remains approximately flat, the interaction is in **saturation** because more context adds no meaningful value. If the frontier falls, the interaction has entered **interference** because requiring more context forces unnecessary or conflicting information into the set.

The frontier makes clear why there is no universal optimal amount of context. One interaction may require a single operational signal. Another may require environment, world, and guardrail context together. What matters is not reaching a target number of signals but reaching adequate coverage with the smallest set.

The frontier is therefore specific to the personalization instance. A change in the customer's current intent can change which provider-side context is relevant, which information is redundant, and where sufficiency occurs. The same provider can consequently have different sufficient sets for different requests.

Testing the frontier requires observing alternative context sets at the same volume. Adding one signal after another along a single path cannot show whether another combination of the same size would have performed better, nor can it fully reveal redundancy, interaction, or interference. This requirement motivates the full-factorial study that follows, in which every combination of the available context signals is observed while customer evidence, current intent, and the generative personalization system are held constant.

## 4 Empirical Test of the Theory

### 4.1 Study Logic and Setting

The empirical study tests the central claim of the theory: **provider-side context should improve personalization when it is relevant to the customer's current intent, add little once the relevant circumstances are already covered, and become harmful when additional context is irrelevant or conflicting.** The core of the experiment is to hold the customer, request, catalog, and generative system constant while changing only the context supplied to the system.

We test the theory using the product catalog of a large home-furnishing retailer in Saudi Arabia. The catalog contains more than 20,000 products with structured attributes. The recommendation system uses generative AI as a translator around the retailer's existing catalog search. First, the model reads the customer evidence, the request, and any provider-side context and decides what to search for. The catalog then returns candidate products. The model subsequently ranks those candidates and returns 25 recommendations. This structure allows us to observe context at two points: whether it changes what the system searches for and whether it changes what the system ultimately recommends.

The main experiment used Qwen3.5-9B, an open-weights generative model. We held the model version, prompts, and generation settings fixed and recorded the random seed for each repetition. Importantly, the model was explicitly instructed to treat context as information rather than as a command, to judge whether each signal was relevant, and not to force irrelevant context into the recommendation. This makes the test conservative. If irrelevant context still affects the output, it does so even when the system has been told to screen such information.

### 4.2 Experimental Design

The experiment crossed five context signals in every possible combination. Each signal could either be present or absent, producing $2^5 = 32$ context conditions, including a no-context baseline and a condition containing all five signals.

We used two constructed customers with different purchase histories and styles. Both made the same request:

"I want to renovate my bedroom and make the space feel more spacious."

Each of the 32 conditions was repeated five times for each customer, using the same five seeds across conditions. The design therefore comprised 320 runs, producing 8,000 ranked recommendation positions. Ten runs failed validation because searches returned too few products; all failures occurred with three or more signals and included the competitor promotion, so their exclusion, if anything, understates the harm of added context. Everything except the context was held constant within each customer: the customer evidence, request, product catalog, prompt, model, and model settings. The full-factorial design is important because the Context-Sufficiency Frontier requires comparison not only across different amounts of context, but also across different sets containing the same number of signals.

### 4.3 Context Signals

The five signals were selected to create a clear test of relevance, redundancy, and interference. They represent three of the five provider-side context types developed in the theory: world, goal, and guardrail context (Table 2). Environment and operational context are not manipulated in this study.

**Table 2.** Context Signals, Types, and Relevance to the Request

| Context signal | Type | Relevance to this request | Appropriate response |
|---|---|---|---|
| Competitor Bedroom Promotion | World | Relevant | Reflect the competitor's bedroom offer, for example through relevant beds or nightstands |
| TikTok Mirror Trend | World | Relevant but partly redundant | Increase attention to mirrors |
| Summer Candle | Guardrail | Irrelevant | Do not change the recommendation merely because of the candle issue |
| Sofa Supplier Bankruptcy | Goal | Irrelevant | Do not introduce sofas simply to reduce stock |
| Wallpaper Inventory Reduction | Goal | Irrelevant and outside the search scope | Do not introduce wallpaper simply to reduce stock |

The authors specified the distinction after an interim inspection of 43 runs and before the complete set of runs was analyzed. The competitor promotion bears directly on a bedroom request. The mirror trend is also relevant, but partly redundant because mirrors already provide a natural response to making a room feel more spacious. The remaining three signals do not help answer the focal request. Sofas and candles do not follow from the request, while wallpaper is additionally outside the scope defined for the recommendation service. All five facts were treated as true for the retailer. What changed across experimental conditions was whether the generative system was given each fact.

This structure gives the theory a demanding test. There are context signals that should help, one that should help only modestly because it overlaps with information already present, and three that the model should ideally ignore. Because only two signals are relevant to the focal request, additional context beyond those signals should provide little additional value and may create saturation or interference.

### 4.4 Measures

We separate a simple change in recommendations from an improvement in recommendations. This distinction is necessary because generative systems can produce different lists across repeated runs even when nothing in the input changes.

Our main outcome is **personalization appropriateness**, measured through the three criteria developed in the theory. **Intent preservation** captures whether the recommended products still serve the bedroom renovation request. **Contextual alignment** captures whether the recommendations reflect the provider-side context that is relevant to that request. **Constraint compliance** captures whether the recommendations remain inside the rules of the service. Overall appropriateness is the mean of these three components.

We also measure **stability**, or how similar recommendations are when the same condition is repeated. Higher overlap across repeated runs means greater stability. Finally, because the system separates search from ranking, we trace each signal through the process. We examine whether it changes the search plan, whether products associated with that signal enter the candidate pool, and whether those products reach the final recommendation. This lets us distinguish **consideration** from **adoption**. A signal can influence what the system considers without ultimately influencing what it recommends.

Because the model varies across identical runs, the frontier must be evaluated relative to that variability. We therefore estimate the best context set of each size using some repetitions and evaluate it on held-out repetitions. The tolerance for declaring one set meaningfully better than another is based on the model's own run-to-run variation. We also examine the marginal contribution of each signal across different starting sets and estimate interactions between signals. The three components were coded by rule from catalog categories. Alignment is the share of the 25 recommendations reflecting a relevant fact (beds or nightstands in neutral colors; mirrors); compliance is the share within the search instruction's scope. The rules were fixed after an interim inspection of 43 runs and before the complete data were analyzed. A binary version of alignment, specified first, was near its ceiling at baseline (.90) and is reported as a robustness check. Intervals come from a bootstrap resampling seeds, and the tolerance for comparing sets (.25) is the 95th percentile of differences between repetitions of the same condition.

### 4.5 Cross-Model Validation

We also conducted a smaller cross-model check to examine whether the main pattern depended entirely on the open-weights model used in the full experiment. A human team selected 50 cases from the experiment and reran the corresponding inputs using GPT-6 Astra. The same directional pattern was observed as in the open-weights model.

We treat this analysis as a robustness check rather than as a second experiment. It does not reproduce all 32 combinations, all repetitions, or the complete frontier. It therefore cannot independently estimate the location of context sufficiency. Its purpose is narrower: to examine whether the central relevance pattern observed in the main experiment also appears when the same personalization problem is processed by a different frontier model.

## 5 Results

Before testing the propositions, we establish the reference point. Personalization appropriateness is the mean of three components: whether the recommendations preserve the customer's current intent, reflect the provider-side context relevant to that intent, and comply with the service constraints. Without provider-side context, intent preservation was .99, contextual alignment was .48, and constraint compliance was 1.00, producing overall appropriateness of .82. The baseline therefore already served the customer's stated request well. The empirical question is whether supplying the right context makes that response more appropriate, and whether supplying context the decision does not need makes it worse. Table 3 summarizes the tests.

**Table 3.** Empirical Tests of the Context-Sufficiency Propositions

| Proposition | Main prediction | Key empirical evidence | Interpretation | Conclusion |
|---|---|---|---|---|
| P1: Relevance | Relevant context improves appropriateness; irrelevant context does not and can reduce it when acted on. | Relevant signals: +.023 (.007, .043). Irrelevant signals: -.034 (-.049, -.019). Competitor: +.029. Mirror: +.012. Candle: -.029. Sofa: -.007. Wallpaper: -.060. | Whether context helped depended on its relevance to the request, not simply on its presence. The model ignored some irrelevant context, considered some, and adopted some. | Supported |
| P2: Sufficiency | A subset of available context can reach the frontier; sufficient sets may differ across personalization instances. | C1: .84 baseline, .94 best smaller set, .88 full set. C2: .81 baseline, .92 best smaller set, .76 full set. Best sets were built around the relevant signals. The best sets did not differ meaningfully across the two customers. | A smaller set was enough to outperform the full context set, but this experiment does not show that different instances require different sufficient sets. | Partly supported |
| P3: Saturation | A signal contributes less when what it adds is already covered by the request or other context. | Competitor signal average contribution: .056; mirror: .012. Competitor contribution was .091 when added to no context, then generally .017 to .025 when added to sets containing one to three signals. | The results are consistent with diminishing value from redundancy. The mirror signal is especially consistent with this logic because the request already implied mirrors, but the design does not isolate redundancy directly. | Limited support |
| P4: Interaction | Complementary signals reinforce one another; conflicting signals weaken one another. | Mirror × competitor: +.026 (.006, .046). Candle × mirror: -.040 (-.060, -.025). Sofa × wallpaper: +.028 (.014, .051). | Context value depends on the other context present. Both positive and negative interactions occurred, although the positive interaction between the two goal signals was not predicted. | Supported with qualification |
| P5: Interference | Beyond sufficiency, irrelevant or provider-goal context reduces appropriateness and stability. | Full set below best smaller set for C1 (.88 vs. .94) and C2 (.76 vs. .92). Irrelevant signals reduced alignment by -.059 and compliance by -.039. Stability fell as context accumulated. Intent preservation changed little. | Interference occurred, but not through the predicted mechanism. Extra context did not pull the response away from the request. It destabilized retrieval, displaced responses called for by relevant context, and sometimes weakened constraint compliance. | Partly supported |

## 5.1 Proposition 1: Relevance

Proposition 1 receives the clearest support. Across the experimental conditions, each relevant context signal increased appropriateness by .023 on the 0 to 1 scale, whereas each irrelevant signal reduced it by .034. The intervals for both pooled effects exclude zero. The result therefore rejects a simple “more context is better” account. What mattered was whether the supplied information helped determine an appropriate response to the customer's current request.

The individual signals show what this means behaviorally. The **Competitor Bedroom Promotion**, which directly concerned the focal bedroom request, increased appropriateness by .029 and increased the presence of relevant beds and nightstands. The **TikTok Mirror** signal strongly increased mirror recommendations, but its improvement in overall appropriateness was smaller, .012, with an interval including zero. The signal was relevant, but much of what it implied was already present in the request to make the bedroom feel more spacious.

The three irrelevant signals produced three different responses from the model. The **Sofa Supplier Bankruptcy** signal was largely ignored and had almost no effect on appropriateness (-.007). The **Summer Candle** signal was sometimes considered during search but almost never adopted in the final recommendations; nevertheless, appropriateness declined by .029 because the signal displaced attention from more useful responses. The **Wallpaper Inventory Reduction** signal was actually adopted and produced the largest decline in appropriateness (-.060). Wallpaper and wall-cladding products entered search, the candidate pool, and the final recommendations despite the service scope excluding such products (Figure 7).

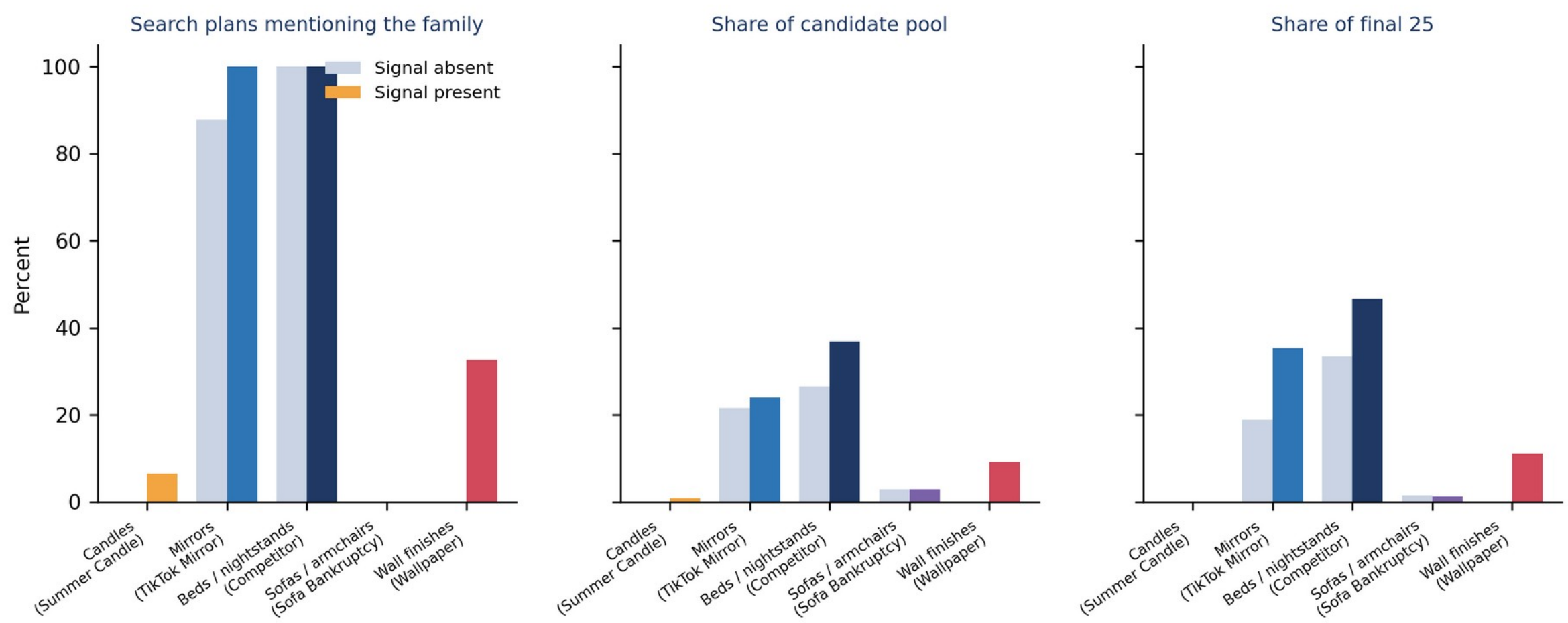

**Figure 7. How Irrelevant Context Enters the Recommendation Process.** Consideration and adoption of the three context signals classified as irrelevant to the focal request. The figure traces each signal from search planning to the candidate pool and final recommendations. Sofa context is largely ignored, candle context is sometimes considered during search but screened out before recommendation, and wallpaper context enters both retrieval and the final recommendation.

The substantive result is therefore stronger than simply saying that irrelevant context is harmful. A generative system can respond to irrelevant context in at least three ways: **ignore it, consider it but screen it out, or adopt it**. Only the first response is harmless. This pattern supports Proposition 1.

### 5.2 Proposition 2: Sufficiency

Proposition 2 asks whether the system needs all available context to reach its best attainable response. The results indicate that it does not.

For Customer 1, held-out appropriateness increased from .84 without context to .94 with a smaller context set, then declined to .88 when all five signals were supplied (Figure 6). For Customer 2, appropriateness increased from .81 to .92 with a smaller set and declined to .76 with the full set. At the smaller context sizes, the best-performing sets began with the signals identified in advance as relevant to the request: the competitor promotion first, followed by the mirror signal.

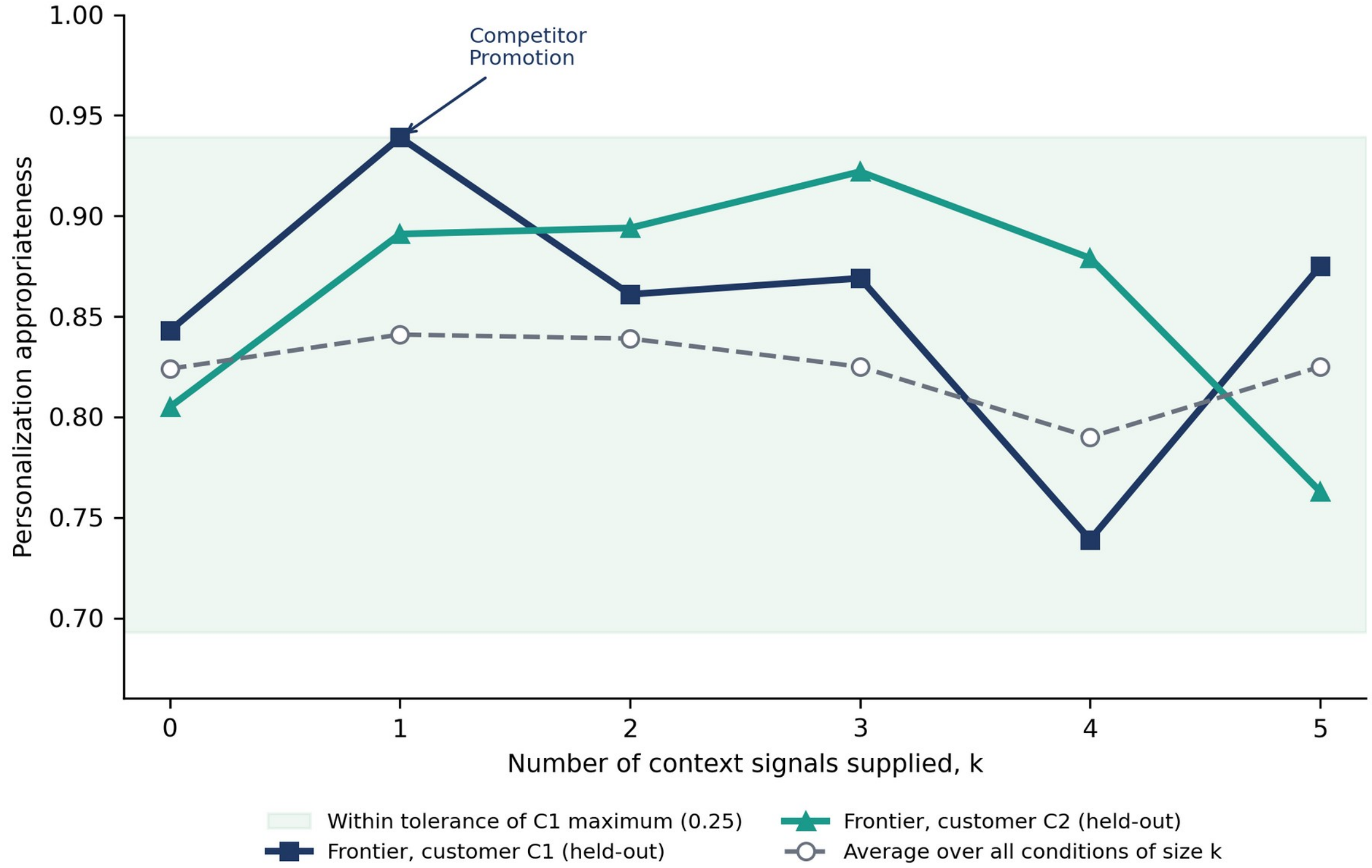

**Figure 6. The Observed Context-Sufficiency Frontier.** Held-out personalization appropriateness for the best context set at each context volume, shown separately for the two customers. For each number of context signals, the best-performing set is identified using a subset of repeated runs and evaluated on held-out repetitions. Appropriateness rises when relevant context is introduced and declines when additional context is forced into the set. The figure illustrates the Context-Sufficiency Frontier but does not identify a precise sufficient volume because differences among nearby points are within the generative system's run-to-run variability.

This pattern is consistent with context sufficiency, but differences between nearby frontier points lie within the model's run-to-run variability; the pooled relevance contrasts under Proposition 1 provide the statistical evidence. **The full set of available context was not the best set for either customer.** More information was therefore not required to obtain the most appropriate response.

The evidence does not, however, identify a precise universal cutoff. Differences among nearby frontier points fall within the model's own run-to-run variability. Nor did the experiment show that the composition of the sufficient set differed across the two customers. The customers differed in history, but they made the same request, and the best sets were similar. Proposition 2 is therefore partly supported: the evidence supports the existence of a smaller sufficient subset, but not the proposition's stronger claim that sufficient-set composition differs across instances.

### 5.3 Proposition 3: Saturation

Proposition 3 predicts diminishing contribution when additional context repeats information already covered by the request or by other context.

The evidence is suggestive rather than decisive. Averaged over all possible starting sets, the competitor-promotion signal contributed .056 to appropriateness, whereas the mirror signal contributed only .012. The competitor signal contributed .091 when added to a condition with no context, but generally only .017 to .025 when added to conditions already containing one to three other signals. Its contribution was therefore much larger when little context was available, although the pattern was not perfectly monotonic.

The mirror signal provides the clearest substantive example. Mirrors were already implied by the customer's request to make the bedroom feel more spacious, and both customer histories already contained a mirror. Adding a mirror-trend signal therefore changed what the system recommended but added relatively little to overall appropriateness.

This pattern is consistent with saturation through redundancy, but the experiment does not independently manipulate redundancy while holding relevance constant. The mirror interpretation is therefore evidence **consistent with** Proposition 3, not a test of its mechanism. Proposition 3 receives limited support.

### 5.4 Proposition 4: Context Interaction

Proposition 4 predicts that the contribution of one context signal depends on the other information supplied with it. The factorial design directly tests this prediction because every combination of the five signals is observed.

Three interactions were distinguishable from zero. The mirror trend and competitor promotion had a positive interaction of .026, meaning that together they improved appropriateness more than their separate effects alone would imply. The candle and mirror signals had a negative interaction of -.040, indicating that their joint presence reduced the contribution each otherwise made. The sofa and wallpaper goal signals also interacted positively, +.028, although this relationship was not predicted by the theory.

The result supports the proposition's central claim that context signals cannot always be evaluated independently. The value of a signal depends partly on the context set into which it enters. The complementary relevant signals reinforce one another, and at least one irrelevant-relevant combination produces a negative interaction. We do not offer a post hoc explanation for the positive interaction between the two goal signals.

Proposition 4 is therefore supported with qualification.

### 5.5 Proposition 5: Interference

Proposition 5 receives important but partial support.

The first prediction is supported. Appropriateness did not continue improving as context accumulated. The complete five-signal condition performed worse than the best smaller context set for both customers: .88 versus .94 for Customer 1, and .76 versus .92 for Customer 2. Recommendation stability also deteriorated substantially as context accumulated. Repeated no-context conditions had a mean Jaccard similarity of approximately .23; with four to five signals, similarity was approximately .12 to .14.

The predicted mechanism, however, was not supported. Proposition 5 expected excessive context primarily to weaken **intent preservation**, meaning that provider-side information would pull the response away from what the customer asked for. Intent preservation barely changed.

Instead, each irrelevant signal reduced **contextual alignment** by .059 and **constraint compliance** by .039 on average. Additional context therefore harmed personalization in a different way. Recommendations continued to look broadly relevant to the bedroom request, but the system became worse at acting on the relevant context and less reliable in respecting the service constraints.

The retrieval process explains much of this effect. As more context was supplied, repeated runs produced increasingly different candidate pools. Most of the decline in final-list consistency originated in what the system searched for rather than in how it ranked an otherwise fixed set of candidates. Filter use in searches rose from 12% of runs without context to 84% with five signals, and the overlap of candidate pools between identical runs fell from .56 to .24. Each added signal reduced the overlap of final lists by .030, but by only .003 once candidate-pool overlap was held constant. The wallpaper condition makes the consequence concrete: irrelevant provider-goal context altered search sufficiently for excluded products to enter the candidate pool, and the downstream ranking stages did not remove them.

This finding **refines Proposition 5 rather than simply confirming it**. Interference does not require the generative system to abandon the customer's request. It can occur while intent remains intact, through unstable retrieval, displacement of responses required by relevant context, or failure to preserve provider constraints.

Proposition 5 is therefore partly supported: interference is observed, but its principal empirical pathway differs from the one originally theorized.

The overall empirical pattern is consequently not a simple inverted-U relationship between the amount of context and personalization quality. The stronger result is **selective**: the trajectory depends on which signals are added. Relevant signals move the system toward the frontier; redundant signals add little; and irrelevant or conflicting signals can move the system away from it.

### 5.6 Cross-Model Robustness

A separate cross-model check provides an initial test of whether the central pattern is specific to the open-weights model used in the factorial experiment. A human team selected 50 experimental cases and reran them using GPT-6 Astra. The directional results matched those obtained with the open-weights model.

Because this check uses a subset of the experiment rather than reproducing the complete factorial design, it should not be presented as a second test of all five propositions or as an independent estimate of the Context-Sufficiency Frontier. It is more appropriately treated as a **cross-model robustness check**: the central relevance and interference pattern was observed in a second generative model.

## 6 Discussion

### 6.1 What the Results Show

The experiment gives a simple answer to the question motivating the theory: more provider-side context is not necessarily better personalization. What matters first is whether the context bears on the decision being made.

Relevant context improved appropriateness. Irrelevant context reduced it. Once the useful context was present, supplying additional signals did not produce better recommendations and eventually reduced consistency and control. The complete context set was not the best-performing set for either customer.

The results also sharpen what we mean by context interference. We initially expected excessive context to redirect recommendations away from the customer's request. Instead, the customer's request remained largely intact. The damage occurred elsewhere. Additional context changed what the system searched for, reduced the consistency of the candidate pool, weakened responses to relevant facts, and in one case allowed excluded products into the final list.

This mechanism is consequential because it can be difficult to see from the final response alone. A recommendation may still look broadly suitable for the customer while being less responsive to the situation, less stable across repeated runs, or less compliant with provider rules.

### 6.2 Contributions to Service and Personalization Research

The first contribution is to extend the situational logic of service to the provider side of personalization. Service research recognizes that value depends on the customer's situation of use (Zeithaml et al. 2020). Our theory adds that the appropriateness of a personalized response also depends on the provider's situation at the moment of service.

This becomes especially important as generative AI takes over tasks previously performed by employees or traditional decision systems (Huang and Rust 2018). Frontline employees customize service by reading such situations and adapting to them (Gwinner et al. 2005; Marinova et al. 2017). A human employee may know that a warehouse is disrupted, a competitor has changed its offer, or a product category has become problematic. A generative system does not automatically inherit that situational judgment. The relevant circumstances must reach the system in some form. Provider-side context is therefore not simply additional data. It is how situational knowledge becomes available to the personalization decision.

Our results show what this transfer involves and what is lost when it is incomplete. An adaptive employee performs two tasks at once: knowing the situation and judging which parts of it matter for this customer and this request. In a generative service system, the provider can supply the first, but the model performs the second. The experiment shows three losses when that selection is delegated. Selection becomes uneven: the model ignored one irrelevant signal, considered another, and adopted a third. Rules do not travel with the context: a constraint applied during search was not carried into ranking. And the service becomes less consistent: identical requests produced increasingly different recommendations as more context was supplied. Situational judgment therefore cannot be transferred simply by transferring situational facts; the provider must also transfer the selection, by choosing what the system receives, and the constraints, by enforcing them at every stage.

The second contribution is to separate **customer evidence** from **provider-side context**. Personalization research has largely focused on improving what the provider knows about the customer. Service frameworks already include contextual information at delivery time (Glushko and Nomorosa 2013); our contribution is to specify which context a decision needs and when more stops helping or harms. Our findings do not challenge the value of richer customer evidence. They show that provider-side context follows a different logic.

Customer evidence can become more informative as the provider learns more about the customer. Provider-side context is useful only to the extent that it bears on the decision being made now. A fact can be highly important in one interaction and irrelevant in another. The relevant theoretical question is therefore not how much context the system can receive, but whether the context supplied covers what the present decision requires.

The Context-Sufficiency Frontier formalizes that distinction. It does not ask for the largest context set. It asks for the smallest context set that reaches the best attainable level of appropriateness for the personalization instance. The frontier therefore changes the stopping rule from "add useful information when available" to "add information until the decision is sufficiently covered."

The third contribution concerns the role of generative AI itself. The model was explicitly instructed to judge the relevance of context. It sometimes did so correctly. It ignored the sofa signal. But it also spent search effort on the candle signal and adopted the wallpaper goal despite the scope restriction. Relevance screening occurred, but it was inconsistent.

This suggests that selecting context cannot simply be delegated to the generative model after all available information has been supplied. What the provider chooses to send to the system remains part of personalization design. Frontline studies already document "AI feeding" routines, in which employees document and supply the information GenAI uses (Reinhard et al. 2026); our results suggest that what is fed, not only whether it is fed, shapes the service. The model's ability to reason over context does not remove the need to decide which context it should receive.

### 6.3 Implications for Research on Generative Personalization

The experiment also has implications for how generative personalization should be studied.

First, recommendation change should not be treated as evidence that a context manipulation had an effect. The recommendation lists in our experiment varied substantially even when the same inputs were repeated. Adding a signal changed the list no more than rerunning the same condition. Effects became visible only when we examined whether the products relevant to a particular signal changed and whether appropriateness improved or declined.

Second, repeated runs are necessary when generative models are part of the service process. A single generated recommendation cannot distinguish an effect of context from the system's ordinary stochastic variation (Ouyang et al. 2025). In our data, this variation increased as more context was supplied.

Third, context should be traced through the service process rather than measured only at the final output. The candle signal affected search but almost disappeared from the final recommendations. The wallpaper signal affected search, candidate retrieval, and final ranking. Treating both simply as "context included" would hide this difference. The distinction between consideration and adoption helps reveal how generative systems translate context into action.

These findings also revise the predicted mechanism of interference. Provider goals did not generally replace the customer's intent. Instead, unnecessary context disrupted how the system acted on relevant information and constraints. This suggests a broader interpretation of generative interference: a signal can be harmful not only when it changes the apparent objective of the response, but whenever it changes intermediate processing in a way that reduces appropriateness, reliability, or control.

## 7 Managerial Implications

The practical implication is that firms need to design personalization at two levels. The first is the customer record, which contains customer evidence. The second is what we call the **decision context package**: the smallest set of current provider-side information needed for a particular personalization decision.

The decision context package is not a fixed profile attached to the customer. It is assembled for the interaction. A customer asking for design inspiration may need one package. The same customer requesting immediate delivery may require a different package because stock and fulfillment capacity become decisive.

For managers, the central discipline is to select context by relevance rather than availability. The fact that a firm knows something does not mean a personalization model needs to know it for every interaction. Context should enter the package when it can change what constitutes an appropriate response to the current request and when that information is not already adequately represented by the request or customer evidence.

The results also show why constraints should be enforced throughout the service process. In the wallpaper case, the rule excluding finishing products existed during search planning but was not repeated during later stages. Once wallpaper entered the candidate set, the ranking process could carry it forward. Firms should therefore not assume that a constraint applied at one generative step will automatically govern every later step.

Provider goals may require similar care. A commercial objective such as reducing inventory is legitimate, but our results suggest that feeding such goals directly into the same generative process that interprets customer intent can create unintended effects. One possible architecture is to generate customer-appropriate options first and apply eligible business priorities only within that set. Our experiment does not test that architecture, so we treat it as an implication for design rather than as an established result.

Managers should also monitor more than customer fit. In our experiment, recommendations continued to look broadly related to the customer's request even while stability and rule compliance deteriorated. Evaluating only whether the output "looks relevant" would have missed these failures. Useful monitoring should therefore include appropriateness, consistency across repeated runs, compliance with constraints, and where context enters the retrieval and ranking process.

Finally, firms can estimate their own Context-Sufficiency Frontier. The purpose is not to test every possible piece of company information. It is to identify a manageable set of context signals that could plausibly matter for a class of service decisions, test combinations of those signals across representative requests, and use repeated runs to separate real effects from generative variability. The resulting frontier provides a practical stopping rule: once the relevant circumstances are sufficiently covered, supplying additional context needs a reason.

## 8 Limitations and Future Research

The findings should be interpreted within three boundaries.

First, the study examines personalization appropriateness, not downstream customer or firm outcomes. Appropriateness captures whether a recommendation serves the customer's current intent, reflects relevant provider-side context, and respects applicable constraints. This allows us to test the theory directly, but it does not establish whether more appropriate recommendations increase conversion, satisfaction, retention, or profitability. Future research can connect the Context-Sufficiency Frontier to these outcomes and examine whether the context set that maximizes appropriateness also maximizes customer or firm value.

Second, the experiment is intentionally controlled. It uses two constructed customer profiles making the same request so that customer evidence, current intent, the catalog, and the generative system can be held constant while provider-side context varies across all 32 combinations. This design permits a clean test of how relevance, redundancy, interaction, and interference operate, but it does not establish how sufficient context sets vary across the broader range of customers, requests, and service settings. The focal request also performed reasonably well without provider-side context, making the observed insufficiency relatively mild. Future studies should vary current intent directly and examine situations in which provider-held information is essential to an appropriate response, such as inventory availability, delivery capacity, eligibility, compatibility, or safety constraints.

Third, the empirical frontier is specific to the systems and context signals tested here. The complete factorial experiment uses one open-weights model, while a smaller 50-case test with GPT-6 Astra reproduces the directional relevance pattern but does not reconstruct the full frontier. The five experimental signals also represent three of the framework's five context types. These choices do not affect the theoretical definition of context sufficiency, but they leave open whether the location and shape of the frontier vary across model architectures and forms of provider-side context. Replications across model families and service domains, particularly with environment and operational context, would establish the extent of that variation.

Several extensions follow from these boundaries. One concerns context freshness. Relevance is not necessarily stable over time: inventory, delivery capacity, competitor actions, market conditions, and operational disruptions can change quickly. Future research should therefore examine not only which context an interaction requires, but how recently that context must have been observed to remain useful.

A second extension concerns system design. The experiment shows that context can affect personalization before the final recommendation is produced because it can alter search and the candidate pool. Additional context also reduced stability and, in some conditions, weakened constraint compliance rather than simply pulling recommendations away from the customer's request. Future research can test whether constraints should be enforced at multiple stages, whether provider goals are better introduced after customer-appropriate options have been identified, and whether generative systems can determine which context they need and request it selectively.

These extensions preserve the central boundary of the paper rather than weakening it: the present study establishes whether the composition of provider-side context matters under controlled conditions. The next question is how the sufficient set changes across customers, decisions, systems, and service environments.

## 9 Conclusion

Personalization has traditionally been framed as a problem of knowing the customer well enough. This paper adds a second requirement: the provider must also know enough about the situation in which the personalized decision is being made.

The two forms of information serve different purposes. Customer evidence helps determine what the customer is likely to value. Provider-side context determines whether a response built on that evidence is appropriate now. Generative AI makes it much easier to bring changing provider-side context into an existing personalization system, but that technical flexibility does not make all available context useful.

The experiment shows why the distinction matters. Relevant context improved personalization. Irrelevant context reduced appropriateness and stability. Additional context could alter retrieval even when the final recommendations remained broadly consistent with the customer's request, and one provider-goal signal carried products through a process that the provider's own rule was intended to exclude. A 50-case cross-model check using GPT Astra 6 produced the same directional pattern, suggesting that the central relevance result is not confined to the open-weights model used for the full experiment.

The theoretical implication is therefore not that personalization systems need more context. They need **sufficient context**. The Context-Sufficiency Frontier identifies the point at which the relevant circumstances of the decision have been adequately covered and additional information stops creating value. The decision context package translates that idea into a managerial unit: provide what this interaction needs, enforce the applicable constraints throughout the process, and do not treat information as useful merely because it is available. For service research, the broader lesson is that moving personalization into generative systems transfers situational judgment only when the provider also transfers its selection.